# The COVID-19 pandemic: socioeconomic and health disparities

Dr Behzad Javaheri

**Abstract**—Disadvantaged groups around the world have suffered and endured higher mortality during the current COVID-19 pandemic. This contrast disparity suggests that socioeconomic and health-related factors may drive inequality in disease outcome. To identify these factors correlated with COVID-19 outcome, country aggregate data provided by the Lancet COVID-19 Commission subjected to correlation analysis. Socioeconomic and health-related variables were used to predict mortality in the top 5 most affected countries using ridge regression and extreme gradient boosting (XGBoost) models. Our data reveal that predictors related to demographics and social disadvantage correlate with COVID-19 mortality per million and that XGBoost performed better than ridge regression. Taken together, our findings suggest that the health consequence of the current pandemic is not just confined to indiscriminate impact of a viral infection but that these preventable effects are amplified based on pre-existing health and socioeconomic inequalities.

## I. INTRODUCTION

The world is facing an unresolved and urgent global health challenge with indications that pharmaceutical interventions alone will not provide a permanent solution to COVID-19 and future pandemics [1]. This crisis is bringing to light and has magnified existing health and socioeconomic inequalities with higher mortality amongst people living in poverty, with pre-existing conditions, unemployed and ethnic minorities. Previous studies have reported that this higher mortality is associated with wealth, social class and ethnicity and that disadvantaged groups are at greater risk of COVID-19 mortality [2-5]. For example, whilst 14% of England and Wales are from Asian and minority ethnic, they nevertheless account for ~35% of critically-affected COVID-19 patients [4].

This disparity in pandemic outcome is not exclusive to the current crisis. Indeed, previous studies reported higher mortality in countries with extreme poverty [6], amongst poor, unemployed and working-class in industrialised countries [7-9] during the 1918 influenza pandemic. More recently, during H1N1 2009 pandemic, higher mortality was reported in disadvantaged groups around the world [10-13].

This historical background suggests that the outcome of COVID-19 has been exacerbated by social and economic determinants of population health. It also implies that, to successfully suppress high mortality, non-pharmaceutical interventions to address these societal inequalities in the present and future health crisis should be considered.

During the current pandemic, Governments around the world have taken non-pharmaceutical steps, implemented social restrictions and provided economic support of various degrees. These include school and workplace closures, restrictions on international travel and test/tracing policies. For Government-led policies to significantly reduce mortality in particular amongst disadvantaged groups, appropriate social, economic and determinant of population health need to be identified and targeted.

It is therefore important to identify socioeconomic and health-related predictors to fine-tune social and economic measures for effective management of the pandemic.

## II. ANALYTICAL QUESTION AND DATA

Fundamental to appropriate management of current health crisis is the identification of factors that are associated with high COVID-19 mortality. This will allow policymakers to provide the needed support to vulnerable communities that bear a disproportionate burden.

One major limitation to ascertain the sources of the disparity is difficulty in obtaining patient-level data where patients individual health and socioeconomic background matched to disease severity and outcome. This arises primarily with concerns surrounding patients data sharing [14]. An alternative approach to partially remedy this limitation is a country aggregate of COVID-19 mortality together with their respective socioeconomic, demographic and health information.

Is it possible to identify these factors, ultimately to address non-viral drivers of mortality? Building on growing evidence and by focusing on the idea that socioeconomic and health background are forerunners for higher COVID-19 mortality, this study will: i) explore the correlation between COVID-19 mortality, societal and health indicators and Governments responses to the crisis and ii) examine whether COVID-19 outcome could be predicted by these indicators alone.

This study aims to address these questions using the Lancet COVID-19 Commission dataset [15]. It currently contains 3.8 million datapoints, 218 countries, 304 days with 135 indicators. These include parameters related to mortality, risk factors including age, gender, and pre-existing conditions (e.g. diabetes), lifestyle (e.g. life expectancy, smoking), healthcare and hygiene provisions (number of hospital beds, handwashing facilities etc.), socioeconomic status (poverty, social mobility, financial support etc.) and government response to the crisis (e.g. closure of public transport, schools and workplaces). These data are pooled from various trusted sources: a) Oxford COVID-19 Government Response Tracker (OxCGRT) [16]; b) Imperial College London YouGov COVID-19 Behaviour Tracker Data Hub [17]; c) Johns Hopkins University [18]; d) Our World in Data [19] and; e) Google COVID-19 Community Mobility Reports [20].

## III. ANALYSIS

To address the aims the following steps are taken:

*A. Data preparation*

Initially, the dimension of imported data [21], and overall structure explored indicating that dataset has 64,675 rows and 138 columns. Describe function in Pandas used to test the overall description of the data, including count, mean, standard deviation, min, max, 25%, 50% and 75%. The count function indicates missing values which was further confirmed by "isnull" and ".isna." test. To impute missing values, columns with significant missing values were dropped. Furthermore, the countries with less than 1,000,000 population were excluded, the analysis focused on top 5 most affected countries (USA, Brazil, India, Mexico and the UK) and confined to data collected between 01/April and 30/October/2020 resulting in a dataset with no missing values (Figure 1).

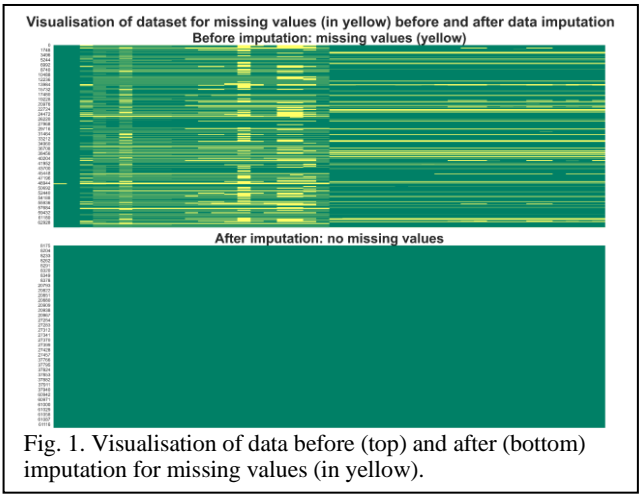

Fig. 1. Visualisation of data before (top) and after (bottom) imputation for missing values (in yellow).

*B. Data derivation*

In this step the following procedures performed: a) feature engineering, b) understanding data by visualisation and c) distribution analysis and data transformation.

Initially, a new feature, "daily recovery/million" computed and visualised with daily mortality suggesting bimodal recovery with the first phase between July-September and second peaking at the end of October (Figure 2).

Analysis of histogram plots reveals that the target variable (first histogram) does not follow a normal distribution (Figure 3). Further dissection using a probability plot to compare the distribution of the target variable to theoretical normal distribution reveals a curvature. To address this, a log transformation was performed; histogram of the transformed variable suggests a bimodal phase for data related to daily COVID-19 mortality potentially due to earlier and current phases of higher mortality (Figure 4).

Additional exploration of select variables (per country) including daily mortality/million, daily recovery/million, measures of Government responses (health containment & stringency policy), socioeconomic including extreme poverty, life expectancy, age and number of hospital beds, reveal consistent daily mortality/million for Brazil, India and Mexico and a wider spread for the UK and USA (Figure 5).

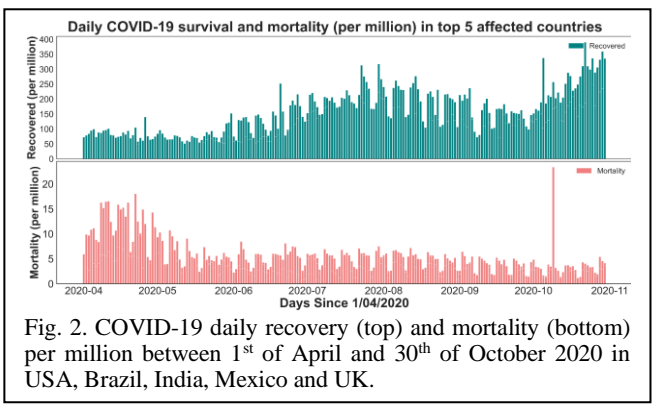

Fig. 2. COVID-19 daily recovery (top) and mortality (bottom) per million between 1st of April and 30th of October 2020 in USA, Brazil, India, Mexico and UK.

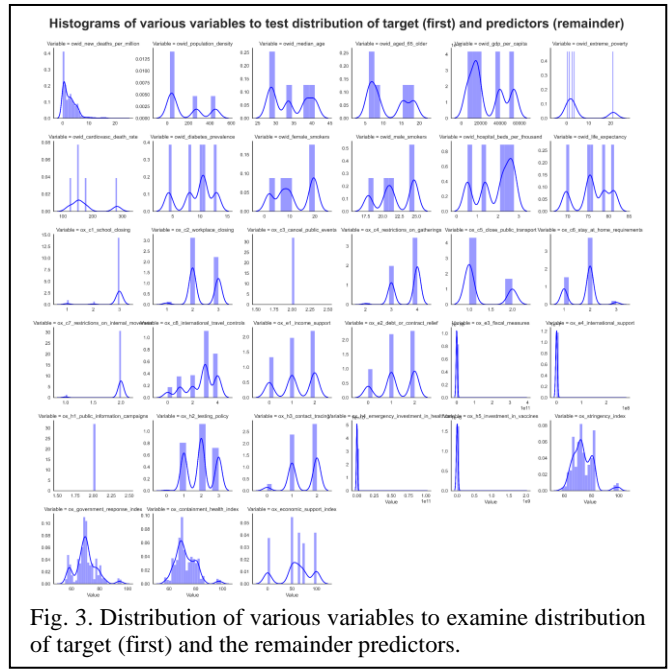

Fig. 3. Distribution of various variables to examine distribution of target (first) and the remainder predictors.

For recovery, a higher variation is observed in Brazil, two clusters for the UK and consistent observation for India and Mexico. Whilst less variation is observed for recovery in the USA than Brazil their average are similar (~115 per million). Furthermore, Extreme poverty is the highest in India and lowest in the UK, with an opposite trend for life expectancy and number of 65 years of age respectively. Number of hospital beds is the lowest in India and highest in the USA and the UK

*C. Construction of models*

In this step following procedures performed: a) correlation analysis, b) testing linearity and multicollinearity and b) construction of ridge regression and extreme gradient boost (XGBoost). The first step will provide some answers on our first question of whether predictors are correlated to mortality. Data show positive and negative correlations of 17 and 13 predictors ("hospital beds/thousand": 0.417; "population density": -0.581) with the target variable. Two predictors ("cancellation of public events" and "public information campaigns") showed no correlation (Table 1).

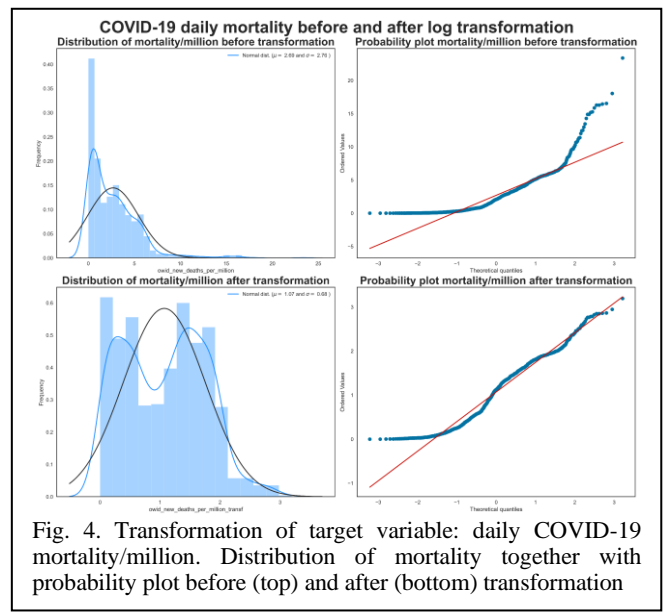

Fig. 4. Transformation of target variable: daily COVID-19 mortality/million. Distribution of mortality together with probability plot before (top) and after (bottom) transformation

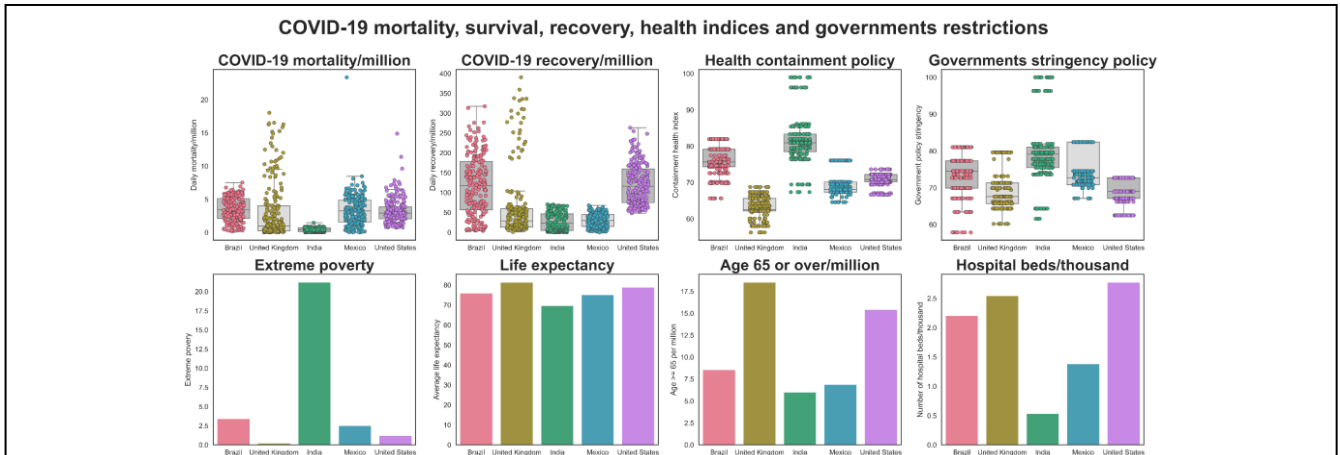

Fig. 5. Country-based visualisation of indices related to COVID mortality, recovery, health, socioeconomics and government restrictions.

Ordinary least square (OLS) testing for linearity showed a violation of linearity suggesting that whilst estimates from OLS are unbiased, however, if employed they will result in high variance (Figure 6). This suggests that data might suffer from multicollinearity. To test for this assumption, the variable inflation factor (VIF) test performed, indicating significant collinearity as evident by high VIF demonstrated in Table 2. Together, these suggest that some predictors are highly dependent on other predictors.

These findings provide evidence to support inclusion of two models: ridge regression as a simple model to deal with collinearity and also able to reduce error by adding a degree of bias (lambda) and extreme gradient boosting (XGBoost) as a model which in addition addresses the relationship between individual predictors and target variable more efficiently.

To construct both models, dataset divided into training and test (80 and 20% respectively). For ridge regression, parameter tuning was performed to identify an appropriate value for penalty lambda following 10-fold cross-validation (CV) using RidgeCV. The purpose of the CV is to identify the best-suited portion of dataset to train the model. Once this is known, the model with the best performance is validated on the entire training dataset and its generalisation tested on the unseen testing dataset.

The average score for CV is shown in Figure 7. The result from cross-validation shows that 0.001 results in the lowest mean square (MSE; metric for error) and highest R2 (metric for performance). This lambda value was used on the entire training datasets and the performance of this model was evaluated on the testing datasets. The R2 values for training and testing datasets were 0.676 and 0.700 suggesting that performance on training and testing was comparable. The plot of residuals and Q-Q plot suggest comparable performance (Figure 7).

To construct XGBoost model, a grid search (GridSearchCV) was performed to identify a number of predictors to select at every split in a given tree, the number of trees, maximum depth of a tree and a learning rate. This resulted in a model with tuned parameters that was subsequently validated on the entire training dataset producing R2 score of 0.845. The R2 score for the testing dataset was comparable (0.82; Figure 8).

| Variable | Score |
|---|---|
| owid_hospital_beds_per_thousand | 0.417 |
| ox_c2_workplace_closing | 0.397 |
| owid_life_expectancy | 0.349 |
| ox_c1_school_closing | 0.308 |
| owid_female_smokers | 0.285 |
| owid_gdp_per_capita | 0.246 |
| owid_median_age | 0.210 |
| owid_aged_65_older | 0.139 |
| ox_c7_restrictions_on_internal_movement | 0.129 |
| ox_c6_stay_at_home_requirements | 0.120 |
| owid_diabetes_prevalence | 0.063 |
| ox_h4_emergency_investment_in_healthcare | 0.057 |
| ox_e1_income_support | 0.052 |
| ox_c4_restrictions_on_gatherings | 0.034 |
| ox_h5_investment_in_vaccines | 0.026 |
| ox_e4_international_support | 0.017 |
| ox_e3_fiscal_measures | 0.008 |
| owid_male_smokers | -0.004 |
| ox_c5_close_public_transport | -0.047 |
| ox_stringency_index | -0.129 |
| ox_h2_testing_policy | -0.178 |
| ox_economic_support_index | -0.236 |
| ox_containment_health_index | -0.259 |
| ox_c8_international_travel_controls | -0.334 |
| ox_government_response_index | -0.370 |
| ox_e2_debt_or_contract_relief | -0.410 |
| owid_cardiovasc_death_rate | -0.450 |
| ox_h3_contact_tracing | -0.472 |
| owid_extreme_poverty | -0.515 |
| owid_population_density | -0.581 |
| ox_c3_cancel_public_events | nan |
| ox_h1_public_information_campaigns | nan |

Table 1. Pairwise Pearson correlation of predictors to target.

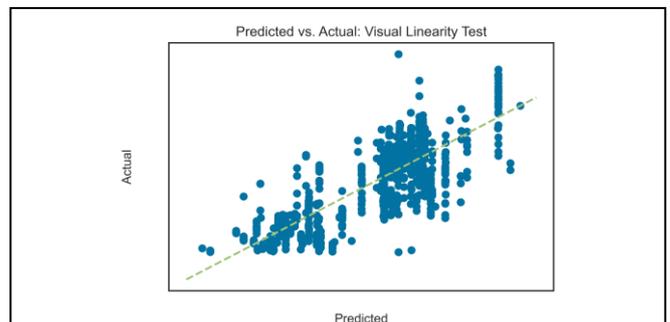

Fig. 6. Ordinary least square testing of linearity.

| | variables | VIF |
|---|---|---|
| 0 | owid_population_density | 3633400264114.962 |
| 1 | owid_median_age | 9007199254740992.000 |
| 2 | owid_aged_65_older | 66719994479562.906 |
| 3 | owid_gdp_per_capita | 818836295885544.750 |

Table. 2. Variable inflation factors (VIF) of selected variables

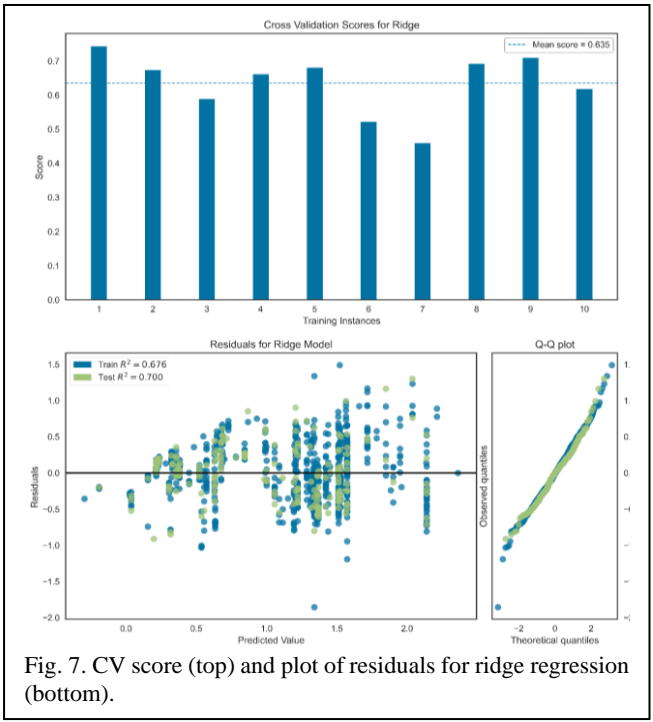

Fig. 7. CV score (top) and plot of residuals for ridge regression (bottom).

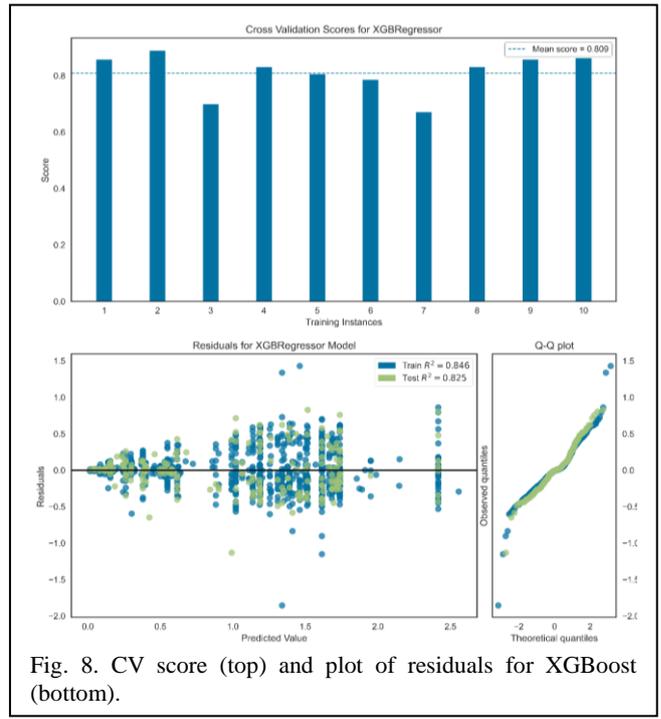

Fig. 8. CV score (top) and plot of residuals for XGBoost (bottom).

*D. Validation of results*

Two steps were taken for validation of results for both ridge regression and XGBoost; a) validation curve and b) plot of prediction error.

Results from validation curve which uses a single hyperparameter on the training and test data suggest that whilst performances of both models are comparable between testing and training, that using lambda as the hyperparameter does not provide us with intuition about the performance of ridge regression (Figure 9). This is a limitation of this model as has a limited capacity for hyperparameter tuning. In contrast, using maximum tree depth for XGBoost suggests that this model performs similarly on the training and testing dataset with various level of tree depth (Figure 10).

Analysis of prediction error which demonstrates actual targets against the predicted values indicate that ridge regression with R2 of 0.7 has higher variance compared with XGBoost with R2 of 0.825 and low variance.

## IV. FINDINGS, REFLECTION AND FUTURE WORK

This study, within its limited scope, strived to address two aims: i) relationship between COVID-19 mortality, societal and health indicators and Governments responses and ii) whether these indicators allowed mortality prediction. Our findings: i) suggest that predictors related to demographics and social disadvantage correlate with daily COVID-19 mortality in the top 5 most affected countries and ii) both ridge regression and XGBoost models revealed that these accounted for 0.7 and 0.83 of the variances for COVID-19 mortality.

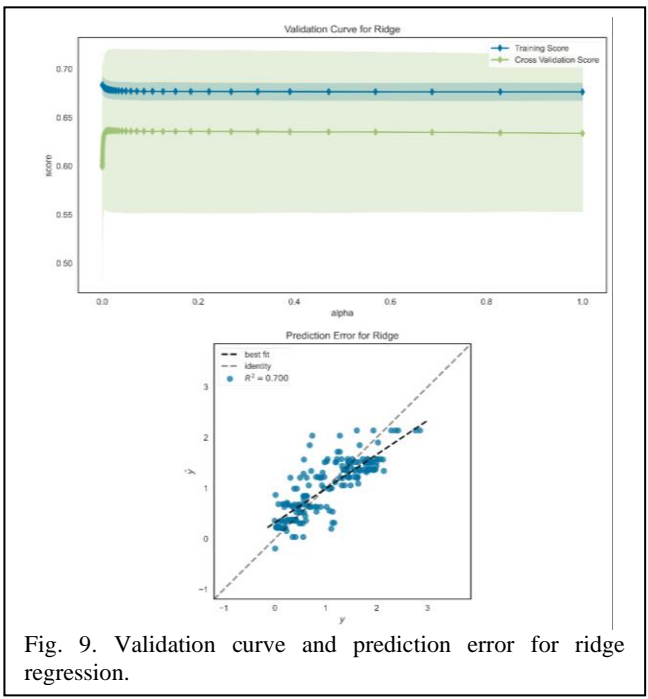

Fig. 9. Validation curve and prediction error for ridge regression.

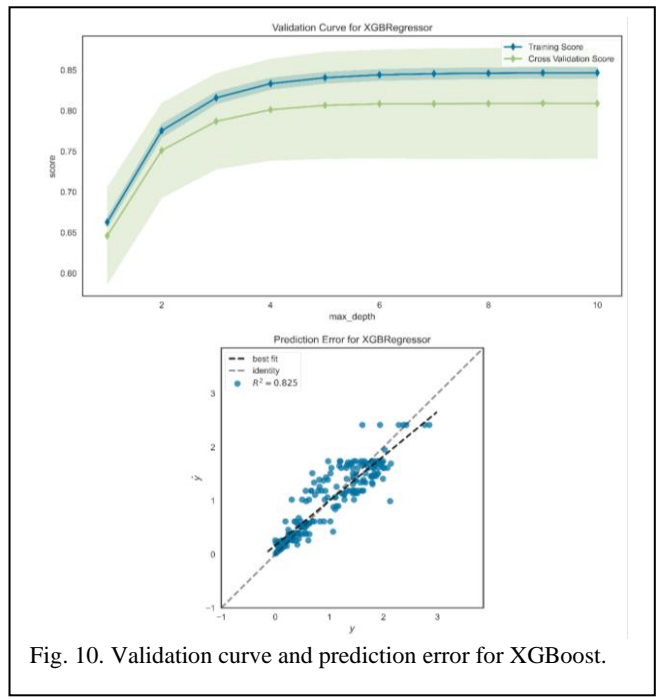

Fig. 10. Validation curve and prediction error for XGBoost.

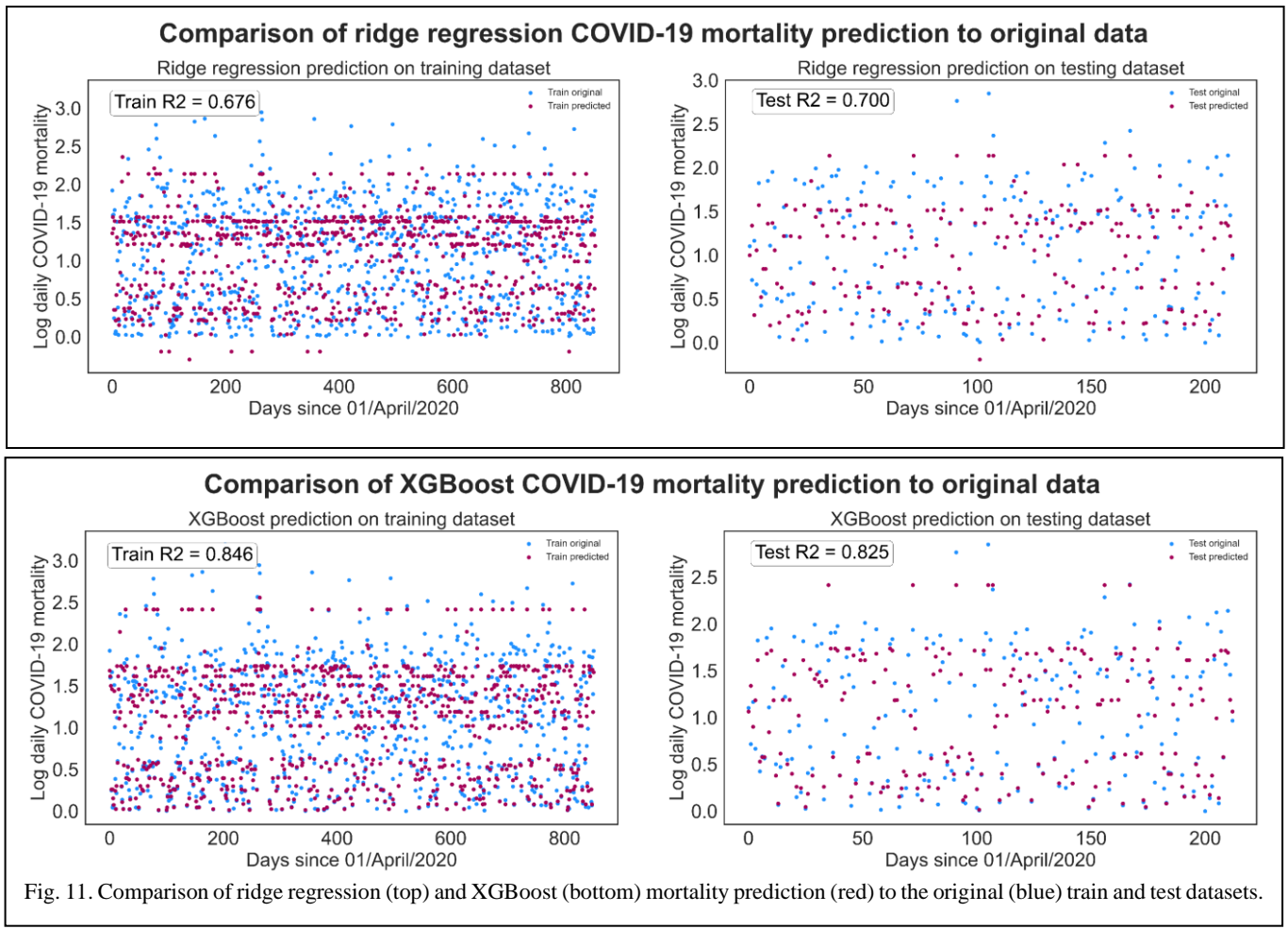

Fig. 11. Comparison of ridge regression (top) and XGBoost (bottom) mortality prediction (red) to the original (blue) train and test datasets.

Performance visualisation of both models (Figure. 11) demonstrate the prediction of COVID-19 outcome with high accuracy. Therefore, health impact of the current pandemic is not just confined to indiscriminate impact of viral infection but that these effects are amplified based on pre-existing health and socioeconomic inequalities and that the outcome of the disease is not just a consequence of infection but a cumulative effect. They also suggest that Government responses of the 5 most-affected countries (USA, Brazil, India, Mexico and the UK) influence the outcome and severity of COVID-19.

Visualisation of mortality and predictors revealed variation and clustering. Whilst time aspect of this dataset was not utilised, this observation, however, might suggest that variations relate to different periods of the pandemic. It is, therefore, reasonable to divide the dataset into periods for better interrogation. This is indeed suggested by Finch *et al.*, (2020) where dividing of data related to different periods of pandemic suggested for dissecting relationship of poverty and mortality in the USA [22]. Additionally, pooling data from countries with similar mortality, government responses as well as the socioeconomic status will allow higher statistical power.

Analysis of predictor importance using better performing model (XGBoost) reveal that indices related to health and socioeconomic status including the number of hospital beds, smoking, cardiovascular death rate, diabetes, age, extreme poverty, GDP as well as those related to Government policies around COVID-19 management including workplace and school closures, test and contact tracing significantly contribute to accurate prediction of disease outcome (Figure. 12).

The relationship between societal and economic indicators and disease outcome in different countries does not seem to follow the same pattern. For example, whilst India has the highest number of people in extreme poverty per million and the lowest number of hospital beds, it has reported lower mortality/million than the UK with the lowest extreme poverty and higher hospital beds. This could potentially be explained by the proportion of the aged 65 years and over per million as the UK has highest such proportion. Indeed, our correlation studies provided evidence to support this notion.

Our findings also suggest that the relationship between health and socioeconomic factors and COVID-19 mortality is complex and that country aggregate data where information related to patients with diverse background combined is not ideal. This is a significant limitation to such analysis [14]. There are several other limitations. The limited scope of this study allowed us to only focus on the top 5 most affected countries; analysis of counties with diverse socioeconomic background might provide a clearer picture. This, however, poses a different challenge as countries with extreme poverty and low GDP lack the capacity to ascertain excess death due to the pandemic. Furthermore, such analysis spanning different countries wrongly assumes that reported statistics adhere to similar standards.

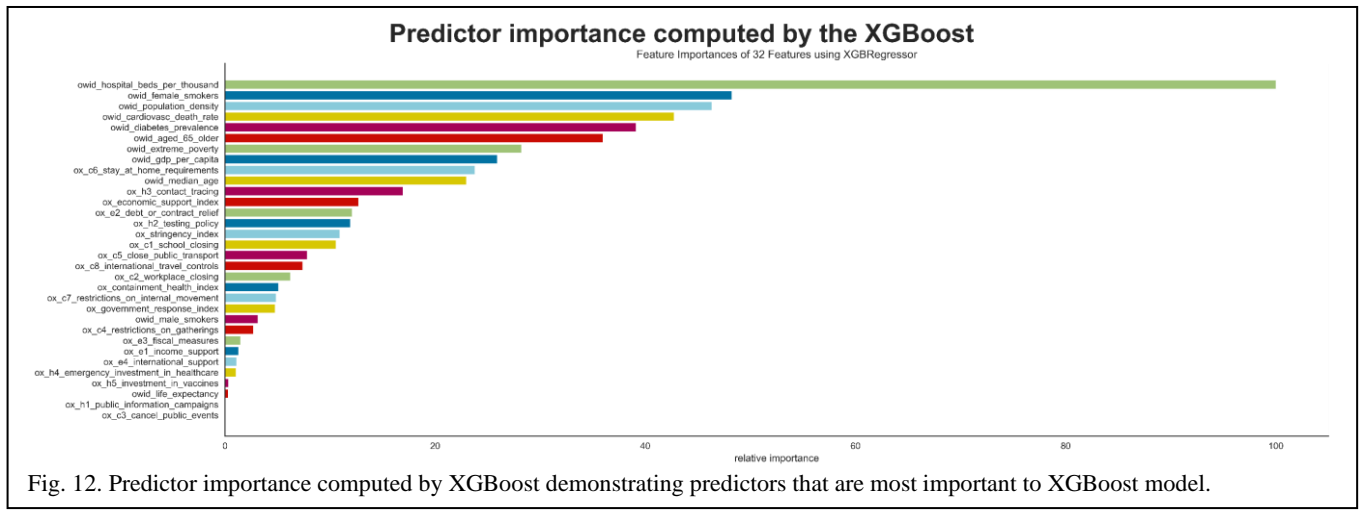

Fig. 12. Predictor importance computed by XGBoost demonstrating predictors that are most important to XGBoost model.

Taken together, this study explored the relationship between a range of variables with COVID-19 mortality and in agreement with historical findings found that variables related to pre-existing health and living conditions correlated with and predicted mortality. Therefore, effective strategies for suppression of the current and future health crisis should consider addressing these societal inequalities.

**Code**

All files related to this work available:

https://github.com/bjavaheri/Data-Science/tree/master/COVID_Lancet_Commission